\documentclass[11pt]{article}

\usepackage[final]{acl}

\usepackage{times}
\usepackage{latexsym}
\usepackage{amsfonts}
\usepackage{amsmath}
\usepackage{booktabs} 
\usepackage{multirow}
\usepackage{subfigure}
\usepackage{xcolor}  
\usepackage[dvipsnames]{xcolor}
\usepackage[table]{xcolor}

\usepackage[T1]{fontenc}
\usepackage[utf8]{inputenc}

\usepackage{microtype}

\usepackage{inconsolata}

\usepackage{graphicx}

\newcommand{\ModelName}{UniCog}
\title{UniCog: Uncovering Cognitive Abilities of LLMs through \\Latent Mind Space Analysis}

\author{Jiayu Liu\textsuperscript{1*}, Yinhe Long\textsuperscript{1*}, Zhenya Huang\textsuperscript{1$\dag$}, Enhong Chen\textsuperscript{1} \\
  \textsuperscript{1}State Key Laboratory of Cognitive Intelligence, 
  University of
Science and Technology of China\\
  \texttt{jy251198@mail.ustc.edu.cn}
  \thanks{Equal Contribution}
\thanks{Correspondence Author}
}

\begin{document}
\maketitle
\begin{abstract}
A growing body of research suggests that the cognitive processes of large language models (LLMs) differ fundamentally from those of humans. However, existing interpretability methods remain limited in explaining how cognitive abilities are engaged during LLM reasoning. In this paper, we propose \textbf{\ModelName}, a unified framework that analyzes LLM cognition via a latent mind space. Formulated as a latent variable model, {\ModelName} encodes diverse abilities from dense model activations into sparse, disentangled latent dimensions. Through extensive analysis on six advanced LLMs, including DeepSeek-V3.2 and GPT-4o, we reveal a Pareto principle of LLM cognition, where a shared reasoning core is complemented by ability-specific signatures. Furthermore, we discover that reasoning failures often manifest as anomalous intensity in latent activations. These findings opens a new paradigm in LLM analysis, providing a cognition grounded view of reasoning dynamics. Finally, leveraging these insights, we introduce a latent-informed candidate prioritization strategy, which improves reasoning performance by up to 7.5\% across challenging benchmarks. Our code is available at \url{https://github.com/milksalute/unicog}
\end{abstract}

\section{Introduction}
Despite their impressive performance on various reasoning tasks, large language models (LLMs) still exhibit striking brittleness in their underlying cognitive abilities. A growing body of evidence shows that even small cognitive perturbations, such as semantic rephrasing or adding distractors, can lead to substantial performance degradation~\cite{li2024gsm,zhu2024dynamic,liucogmath}. In contrast, humans behave differently. Once we grasp a problem, we can reliably generalize to cognitively equivalent variants with little effort~\cite{fodor1988connectionism,lake2023human}.
%This fragility stands in sharp contrast to human cognition: 
This discrepancy raises an important but poorly understood question in LLM research: what cognitive processes do LLMs employ when they reason? More specifically, how do different cognitive abilities, such as problem comprehension, logical deduction, and knowledge application, manifest in the reasoning dynamics of a model? 
%Understanding these aspects is essential for moving beyond benchmark-driven evaluation toward a principled characterization of LLM reasoning competence.
% Such discrepancies point to a deeper issue: we still lack a clear understanding of what cognitive abilities LLMs actually manifest during reasoning. Do they truly comprehend the problem? Do they follow coherent logical steps? Or are they relying on shallow pattern matching that fails under distributional shifts? Addressing these questions is essential for moving beyond benchmark-driven evaluation toward a principled understanding of LLM reasoning.
%These questions represent an important but underexplored direction in LLM research. 

\begin{figure}[t]
\centering
\includegraphics[width=\linewidth]{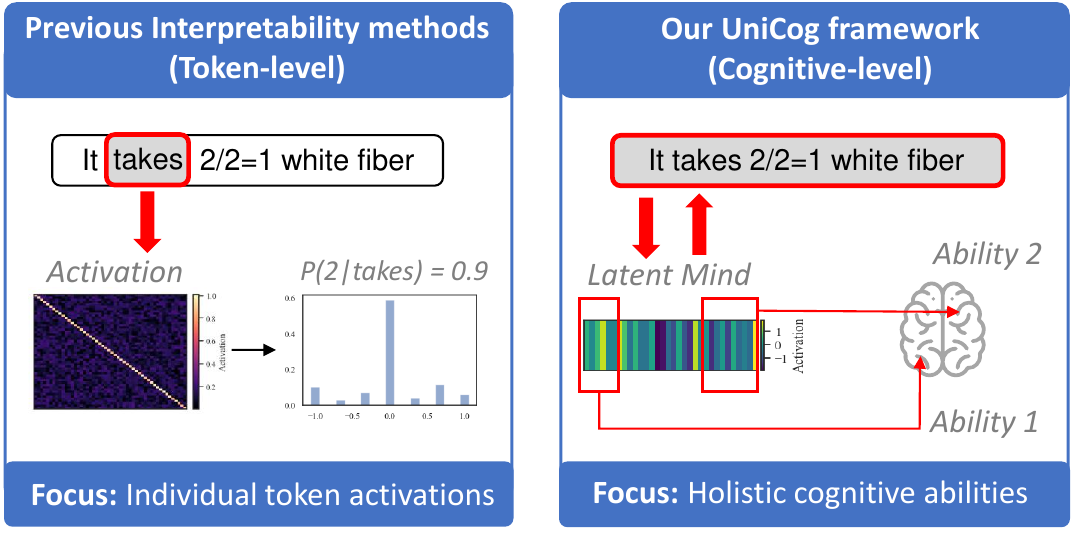}
\caption{Comparison of previous interpretability methods and our {\ModelName} framework.}
\label{figure_intro}
\vspace{-5pt}
\end{figure}

Existing research on model interpretability, such as probing~\cite{belinkov2022probing} and mechanistic interpretability~\cite{conmy2023towards,rai2024practical,rai2024investigation}, primarily focuses on explaining the information stored in individual token activations as shown in Figure~\ref{figure_intro}. While valuable, they fall short of providing a higher-level cognitive perspective on model behavior. For instance, when an LLM fails a mathematical task, the failure may stem from a breakdown in comprehension, logic, or knowledge. Inspecting single activations rarely reveals which cognitive component is responsible. In contrast, such fine-grained diagnostic information is crucial. It not only provides actionable signals for aligning LLM cognition with humans, but also guides which specific abilities should be strengthened when models encounter novel tasks~\cite{zhuang2023efficiently,niu2024large,ivanova2025evaluate}.

In this paper, we propose a general framework that analyzes the cognitive abilities of LLMs. 
%A central challenge is how to represent these heterogeneous abilities. 
Specifically, we unify different abilities into a common continuous space, which we refer to as the latent mind space, and formulate a latent variable model (LVM) over the observed \emph{sentence}-level activations $X$ and the unobserved latent mind $Z$. This design is motivated by pyschological evidence that humans first form internal neural states during thinking and then express them in observable text~\cite{levelt1993speaking,friston2010free}. A key insight of this formulation is that the likelihood $p(X \mid Z)$ can be approximated by standard language modeling.
%thereby sidestepping the need for internal access to closed-source LLMs. 
As for the posterior $p(Z\mid X)$, directly modeling it as a dense mapping often leads to entangled information in $Z$~\cite{esmaeili2019structured}. Inspired by the functional modularity of the human brain
%, where different cortical areas specialize in different cognitive functions
~\cite{kanwisher1997fusiform,huth2016natural}, we argue that an effective latent mind should exhibit sparse structure, such that only a subset of latent dimensions is responsible for each reasoning ability. To this end, we introduce a sparse mapping in the posterior model, which enforces dimension-wise sparsity in the latent mind and across different reasoning abilities~\cite{gaoscaling}.

% The entire model is trained by maximizing the Evidence Lower Bound (ELBO), allowing it to be learned directly from existing reasoning data without any additional data synthesis. More importantly, it provides a unified, architecture-agnostic latent space for $Z$. This establishes a single probabilistic framework for evaluating reasoning explanations across heterogeneous model architectures.

We conduct a systematic empirical study on six state-of-the-art LLMs. Our experiments demonstrate that:  
(1) the latent mind space of LLMs follow a Pareto principle, where a shared reasoning core is complemented by sparse, ability-specific cognitive signatures; (2) LLMs' latent minds can capture patterns for different cognitive variants; and (3) there is a cognitive amplification effect, where difficult variants will lead to intensified latent activations, typically 1.1$\times$ to 2.0$\times$ stronger.

Furthermore, we discover that the latent minds explicitly encode reasoning correctness, allowing for a fine-grained analysis of model behavior. Leveraging these insights, we propose a plug-and-play prioritization strategy to guide LLMs toward reliable reasoning paths. This approach achieves an accuracy gain of up to 7.5\% across four challenging benchmarks. Together, our work suggests that latent mind provides a principled bridge between model cognition and reasoning performance, opening a new paradigm in LLM analysis that offers a cognitively grounded view of reasoning dynamics.

\section{Related Work}
\subsection{Cognitive Abilities of LLMs}
Recent LLMs have demonstrated remarkable cognitive abilities across a wide range of tasks, such as language understanding, numerical reasoning, and logical inference~\cite{wei2022chain,zhao2023survey,ivanova2025evaluate,huang2025foundation}. 

However, a growing body of work has highlighted the brittleness of these abilities. Even minor perturbations on existing benchmarks can cause significant drops in model performance~\cite{li2024gsm,zhu2024dynamic,liucogmath,kargupta2025cognitive,zhao2025coreeval}. For example, Li et al.~\shortcite{li2024gsm} constructed eight variants targeting abilities such as numerical processing and critical thinking, which showed that 25 LLMs suffer substantial robustness degradation. Similarly, Liu et al.~\shortcite{liucogmath} evaluated LLMs along nine cognitive dimensions spanning comprehension, numerical reasoning, knowledge application, and reflective thinking, revealing that different LLMs have uneven cognitive strengths and weaknesses.

While these studies provide valuable assessments of LLM cognitive abilities, they primarily focus on external evaluation. What remains underexplored is how the internal behavior of LLMs relates to these cognitive abilities, and how different reasoning capacities are reflected within the model’s underlying processes.

\subsection{Interpretability of LLMs}
Following Zou et al.~\shortcite{zou2023representation} and Rai et al.~\shortcite{,rai2024practical}, research on interpreting LLMs can be categorized into bottom-up and top-down approaches.

Bottom-up approaches reverse engineer the internal processes of neural networks and analyze their most basic functional units, such as individual neurons. The representative direction is mechanistic interpretability~\cite{conmy2023towards,rai2024practical,geiger2025causal}, which attempts to reconstruct circuits underlying model computations.

\begin{figure*}[t]
\centering
\includegraphics[width=\linewidth]{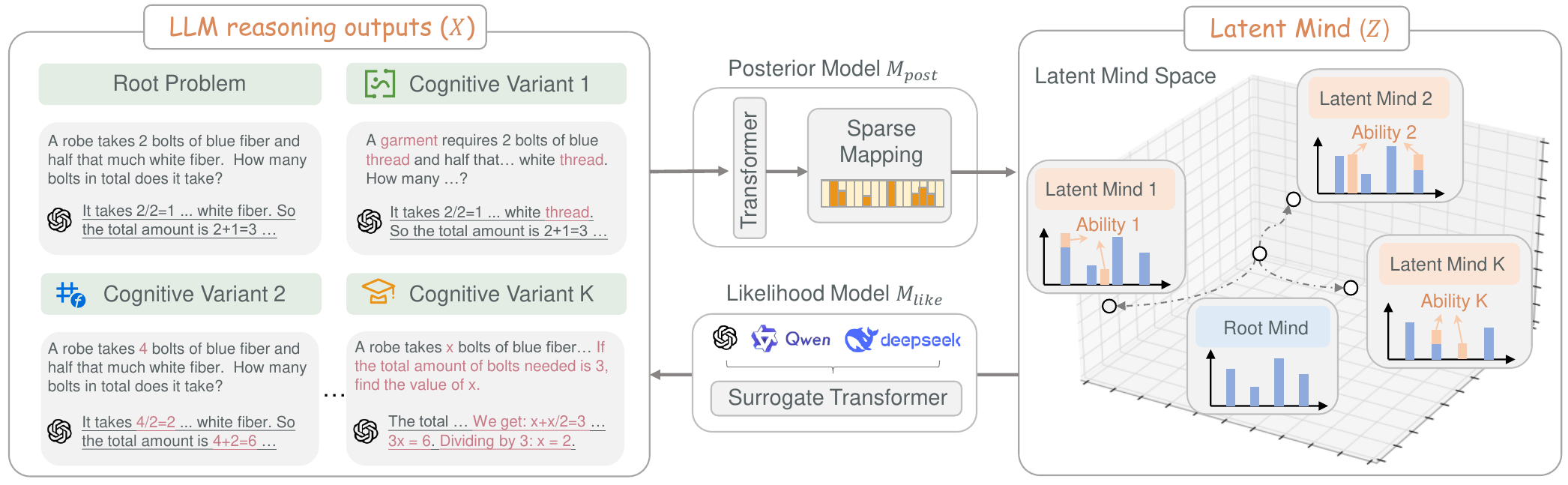}
\caption{Illustration of our {\ModelName} framework.}
\label{figure_framework}
\vspace{-15pt}
\end{figure*}

In contrast, top-down approaches begin with observable model behaviors and seek explanations at the level of representations. This paradigm includes techniques like saliency maps~\cite{simonyan2013deep}, 
%which highlight input regions the network attends to, 
and probing~\cite{belinkov2022probing,ju2024large,li2024llms}, 
%which trains linear classifiers to predict concepts from intermediate representations. 
A more advanced direction is representation engineering (RepE)~\cite{zou2023representation}, which extends this work by manipulating internal representations to steer model behavior. Our work falls into this category, but differs in its focus. We investigate how internal model representations correspond to distinct \emph{cognitive abilities}. By introducing a latent mind space, we establish a systematic connection between different cognitive abilities and interpretable dimensions.

\section{Our Framework}
Our goal is to analyze the cognitive abilities of LLMs. Specifically, psychological studies suggest that during the process of thinking, humans first form internal neural states in the brain that integrate perception, memory, and intention, and then translate these states into observable text~\cite{levelt1993speaking,friston2010free}. This internal state reflects a joint blending of multiple cognitive abilities, which compresses various reasoning and knowledge processes~\cite{fauconnier2008way}.  

Motivated by this view, we unify the diverse abilities of LLMs into a single continuous representation space, which we refer to as the latent mind space. In this formulation, each cognitive ability is characterized and invoked by a distinct activation pattern over a subset of latent dimensions.

Based on this perspective, we propose \textbf{{\ModelName}}, a unified analysis framework for LLM cognitive abilities as shown in Figure~\ref{figure_framework}. Within this framework, our objective is twofold: (1) infer the latent mind directly from the model’s output, and (2) characterize how individual dimensions of this space relate to concrete cognitive abilities. To operationalize this, we adopt a latent variable formulation in which the latent mind serves as the hidden state governing observable outputs.

% To this end, we unify these heterogeneous abilities can be unified and represented within a shared continuous space, which we refer to as the latent mind. This space serves as an abstract representation of the internal cognitive factors underlying an LLM’s reasoning process.

% We aim to infer this latent mind directly from the model’s outputs and further uncover how different dimensions of the latent mind correspond to distinct reasoning abilities. To support this objective, we formalize the problem using a latent variable framework, where the latent mind acts as a structured intermediate representation that governs the generation of observable text.

% Our goal is to uncover the latent mind that governs the thinking process of LLMs. We formalize this objective through a latent variable framework.

\subsection{Latent Variable Formulation}

Let $X$ denote the observed LLM activations of an output sentence (e.g., a mathematical solution), and $Z \in  \mathbb{R}^d$ denote the latent mind, which is a vector in $d$-dimensional space. We assume the following generative process:
\begin{equation}
Z \sim p(Z), \quad X \sim p_\theta(X|Z),
\end{equation}
where $p(Z)$ is a prior distribution and $p_\theta(X|Z)$ is the likelihood that simulates the human process of translating latent minds into observable results. 

\begin{figure}[t]
\centering
\includegraphics[width=\linewidth]{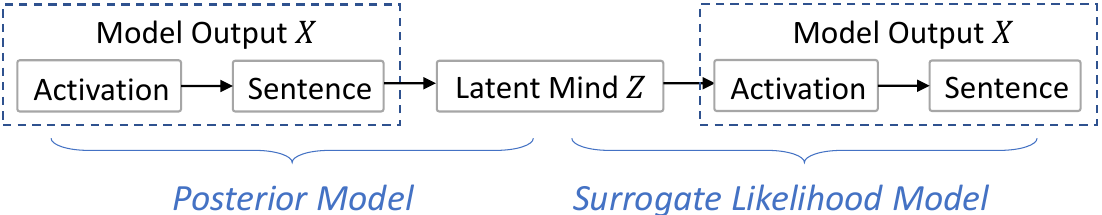}
\caption{Surrogate likelihood and posterior modeling based on language space.}
\label{figure_likelihood}
\vspace{-10pt}
\end{figure}
However, directly modeling the likelihood of activations is intractable, because the transition between different activations relies on the discretization of tokens. In contrast, LLMs are autoregressive token generators, which naturally defines a likelihood in the language space. Therefore, we propose to use language as a bridge, approximating the activation dynamics through the lens of language generation, as shown in Figure~\ref{figure_likelihood}. 
%Crucially, as LLMs are natively autoregressive token predictors, there exists no ground-truth likelihood $p_\theta(X|Z)$. To address this, 
Specifically, we employ a \textbf{surrogate likelihood model} $\mathcal{M}_{\text{like}}$, which reformulates the likelihood as a conditional generation of the language corresponding to specific activations. 
%to distill $p_\theta(X|Z)$ from multiple LLMs simultaneously. 
As illustrated in Section~\ref{section:finding}, this model can faithfully reproduce the output sentences and token distributions of the original LLMs, allowing its internal activations to serve as a reliable proxy. In the following, we do not distinguish between the activations and their resultant output sentences, referring to both as outputs $X$ for brevity.

% Moreover, this design makes our framework applicable to both open-source and closed-source LLMs. From the perspective of the latent mind space, this distillation can be further interpreted as: with a sufficiently large dimension $d$, the latent minds of different LLMs can be embedded and unified within a common space. 
%of It is important to note that $p_\theta(X \mid Z)$ does not represent the forward pass . Instead, we hypothesize that by setting the latent dimension $d$ large enough, the latent minds of different LLMs can be further unified within a common high-dimensional space. Therefore, we can use a single model, which we call the, even for closed-source LLMs. %Furthermore, since the space of $X$ is complex, we implement $\mathcal{M}_{\text{like}}$ can be implemented as language modeling, which will be detailed in Section X.X.
%$p(Z|X)$. Since the true posterior $p(Z|X)$ is intractable, we introduce a variational approximation 

The posterior $q_\phi(Z|X)$ is parameterized by another \textbf{posterior model} $\mathcal{M}_{\text{post}}$ with parameters $\phi$. To optimize both models, we maximize the Evidence Lower Bound (ELBO) $\mathcal{L}_{\text{ELBO}} =$
\begin{equation} 
\mathbb{E}_{q_\phi(Z|X)}[\log p_\theta(X|Z)] 
-\mathrm{KL}(q_\phi(Z|X) \| p(Z)).
\end{equation}

% This objective encourages the latent variable $Z$ to capture all necessary information for reconstructing the solution while remaining compact and close to the prior.

\subsection{Transformer-Based Parameterization}
Now we discuss how to implement the two models in our framework. As discussed above, we implement $\mathcal{M}_{\text{like}}$ as a conditional language model. Specifically, $\mathcal{M}_{\text{like}}$ takes the latent mind $Z$ as a prefix-conditioned context and generates the output $X$ autoregressively. In practice, we instantiate $\mathcal{M}_{\text{like}}$ with a Transformer.
\begin{equation}
\label{eq_likelihood}
p_\theta(X \mid Z) = \prod_{t=1}^{T} p_\theta(x_t \mid x_{<t}, Z).
\end{equation}

The posterior model $\mathcal{M}_{\text{post}}$ takes the token sequence $X$ as a whole to infer the latent mind $Z$. Therefore, we also implement $\mathcal{M}_{\text{post}}$ as a Transformer and parameterize a Gaussian posterior:
\begin{equation}
\begin{gathered}
    \mu(X) = \text{Transformer}(X), \\
    q_\phi(Z \mid X) = \mathcal{N}(\mu(X), \sigma^2 I).
\end{gathered}
\end{equation}
where $\sigma$ is a hyperparameter. However, representing $Z$ as a dense vector often leads to highly entangled information, which hinders fine-grained analysis. To achieve structural clarity, we draw an analogy to the modular organization of the human brain, where distinct cognitive functions are localized to specialized regions~\cite{kanwisher1997fusiform,huth2016natural}. Similarly, we aim to ensure that distinct abilities in $Z$, such as dimensions controlling semantic understanding versus logical deduction, are functionally separated.

Specifically, we leverage the idea of sparse autoencoders (SAE) and introduce a lightweight sparse mapping on top of the Transformer:
\begin{equation}
\mu(X) = \text{ReLU}(\mathbf{W}_1 \cdot \text{Transformer}(X) + \mathbf{b}_1).
\end{equation}

Instead of relying on an $\ell_1$-regularized objective, we empirically find that a k-sparse formulation yields better disentanglement. Therefore, we retain only the top-$K$ activations in $\mu(X)$.

\section{Analysis on Cognitive Abilities of LLMs}

In this section, we analyze the cognitive abilities of different LLMs based on our {\ModelName} framework. 
%Rather than treating reasoning as a black-box input–output process, we aim to uncover how different cognitive abilities are manifested and evolve internally during the reasoning process.
\subsection{Experimental Setup}
\paragraph{Framework Training.}
%To enable a powerful reconstruction from latent minds back to language space, 
We train our framework on NuminaMath-CoT dataset~\cite{numina_math_datasets}, which contains approximately 860K mathematical reasoning samples. This large-scale dataset provides diverse reasoning trajectories and ensures that the surrogate model $\mathcal{M}_{\text{like}}$ can approximate the likelihood of various LLMs (will be verified below). Notably, to facilitate fine-grained analysis of the evolution of latent mind during reasoning, we segment reasoning samples at the sentence level. Training details are provided in Appendix~\ref{append:implement}.
%This design allows the model to align latent representations with intermediate reasoning steps, rather than only final answers.
\begin{table}[t]
\centering
\caption{Definitions of Cognitive Variants on CogMath.}
\label{tab:cognitive_variant_mapping}
\small
\begin{tabular}{ll}
\hline
 & \textbf{Definition} \\
\hline
Cognitive Variant 1 & Sentence Paraphrasing \\
Cognitive Variant 2 & Sentence Disruption \\
Cognitive Variant 3 & Missing Condition \\
Cognitive Variant 4 & Redundant Condition \\
Cognitive Variant 5 & Analogical Reasoning \\
Cognitive Variant 6 & Numerical Transformation \\
Cognitive Variant 7 & Knowledge Redefinition \\
Cognitive Variant 8 & Intermediate Step Questioning \\
Cognitive Variant 9 & Backward Reasoning \\
\hline
\end{tabular}
\vspace{-4pt}
\end{table}
\paragraph{Evaluation Benchmark.}
We analyze the reasoning outputs of LLMs on CogMath benchmark~\cite{liucogmath}, which is constructed by rewriting some \emph{root problems} into nine cognitive variants summarized in Table~\ref{tab:cognitive_variant_mapping}. 
%For example, in Figure~\ref{figure_framework}, given the root problem ``A robe takes \dots'', variant~1 reformulates the problem into an equivalent version without altering the reasoning logic (e.g., by changing the object from ``robe'' to ``garment''). 
For simplicity of analysis, we treat the ability required to solve each variant as a distinct cognitive ability. We acknowledge that each variant may in fact involve multiple abilities. This simplification is used to clearly distinguish and discuss these variants in the following sections.
%Each root problem is rewritten into nine variants, where each variant emphasizes a specific cognitive ability as summarized in Table~\ref{tab:cognitive_variant_mapping}. 

We analyze six representative LLMs, covering both closed-source and open-source settings:
DeepSeek-V3.2-Thinking~\cite{liu2025deepseek}, GPT-4o-0513~\cite{openai_gpt4o}, GPT-3.5-Turbo~\cite{chatgpt2023}, Qwen3-8B~\cite{yang2025qwen3}, LLaMA3.1-8B-Instruct~\cite{grattafiori2024llama}, and Mixtral-8x7B-Instruct~\cite{mistral2023mixtral}.

In Table~\ref{tab:cognitive_variant_mapping}, except for the eighth variant, all other rewritten problems have standard ground-truth answers, enabling us to label each model’s reasoning result as correct or incorrect. This property allows us to analyze and compare the latent minds under both successful and failed reasoning cases. Consequently, our subsequent analysis focuses on the remaining eight cognitive abilities.

\begin{figure*}[t]
\centering
\subfigure[]{
    \includegraphics[width=0.34\linewidth]{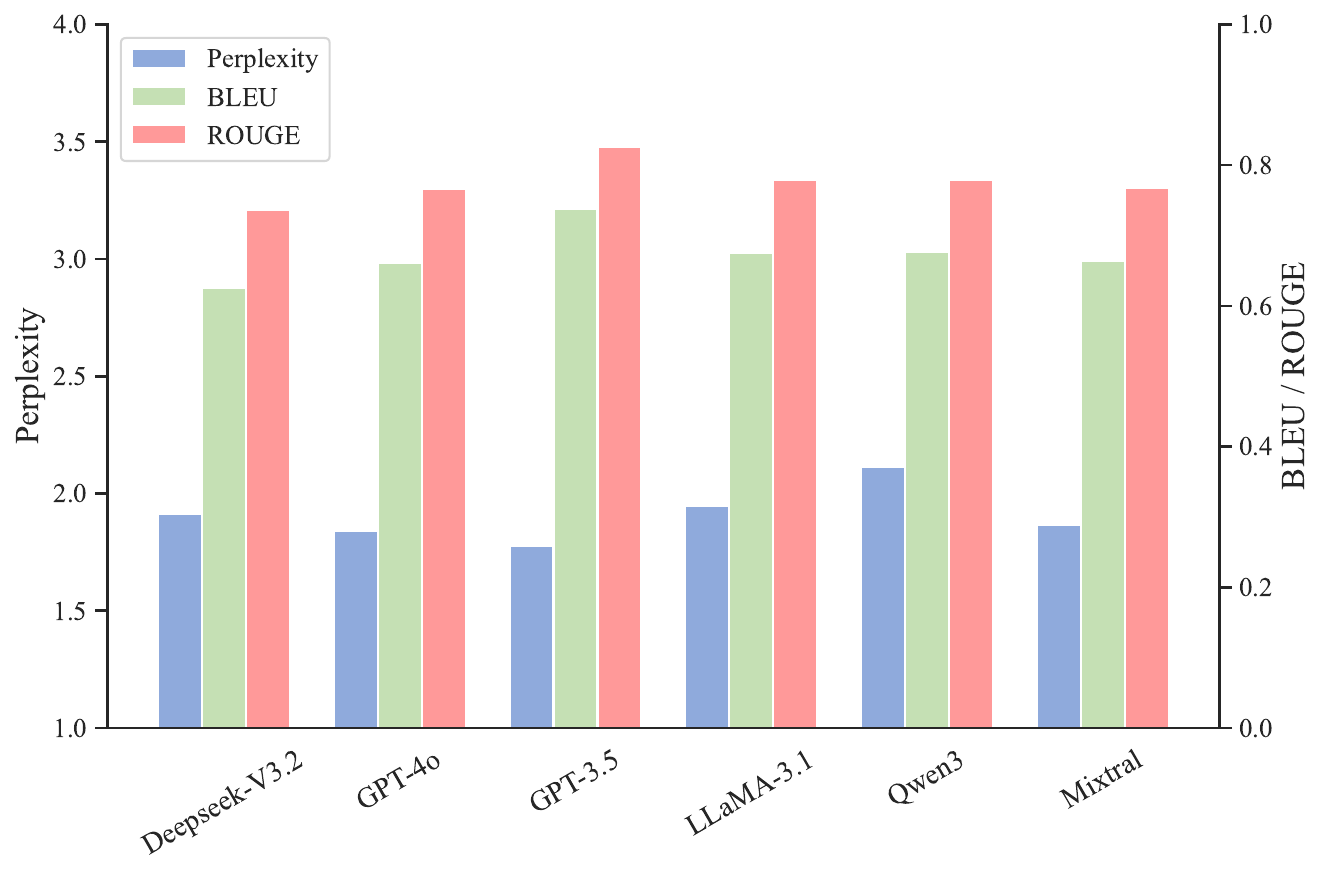}
}
\hfill
\subfigure[]{
    \includegraphics[width=0.32\linewidth]{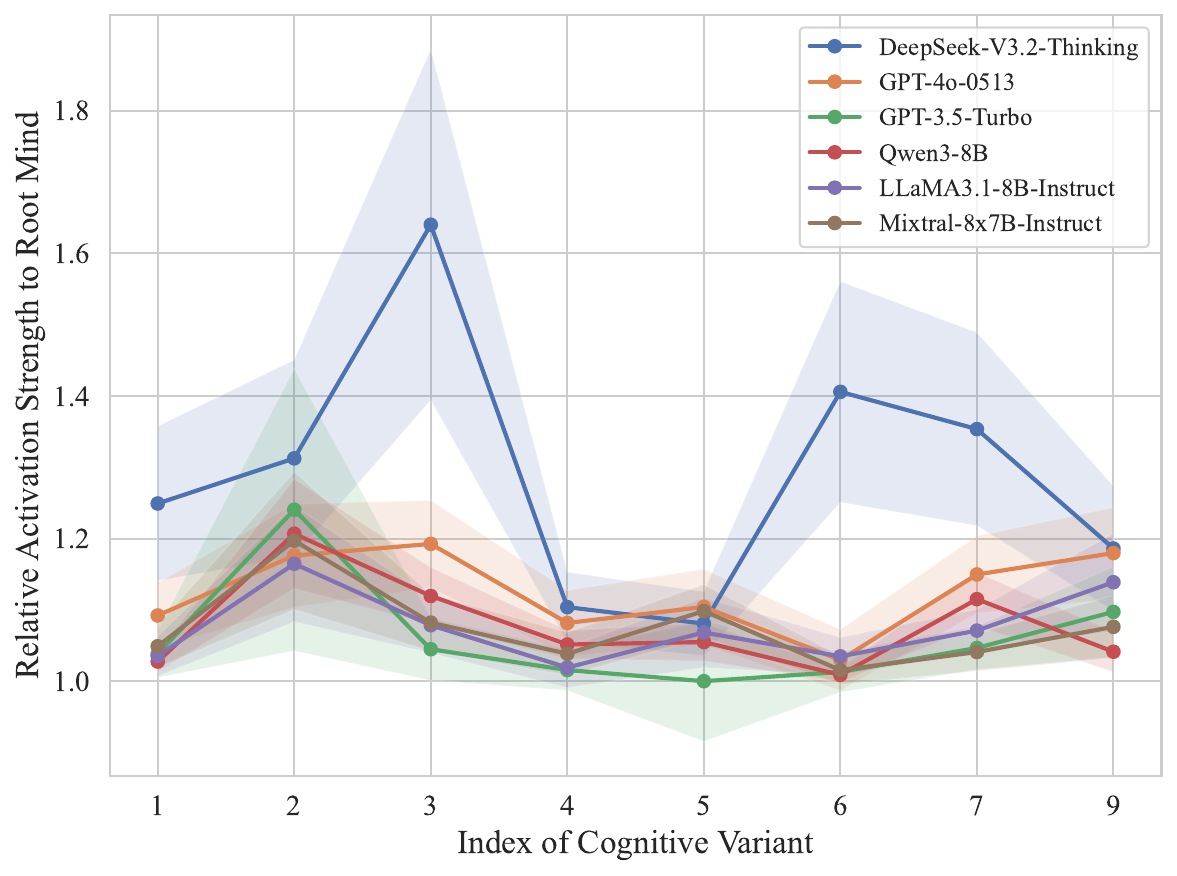}
}
\hfill
\subfigure[]{
    \includegraphics[width=0.28\linewidth]{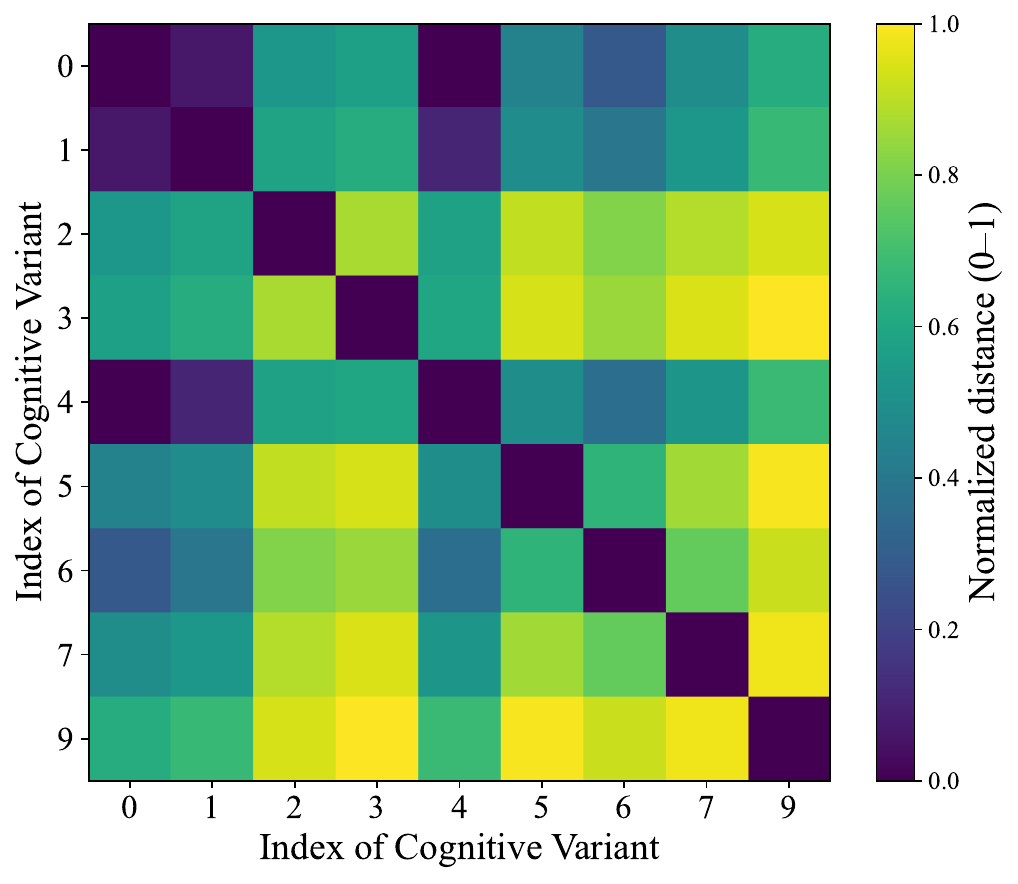}
}
\vspace{-6pt}
\caption{
(a) Perplexity of reasoning outputs sampled from LLMs evaluated by our surrogate model (left y-axis). BLEU/ROUGE scores of reconstructed outputs relative to the original outputs (right y-axis). (b) Average activation strength relative to root mind for each cognitive variant.
(c) Normalized distances among cognitive variants.
}
\label{fig:combined_analysis}
\vspace{-12pt}
\end{figure*}
\subsection{Our Findings}\label{section:finding}
\paragraph{\textbf{Our surrogate model $\mathcal{M}_{\text{like}}$ can accurately simulate the generation of diverse LLMs.}}
We first validate the effectiveness of our surrogate likelihood model by evaluating its reconstruction ability on the outputs of different LLMs. As shown in Figure~\ref{fig:combined_analysis}(a), the average perplexity of their reasoning responses on CogMath, as computed by our surrogate model, remain consistently below $2$. Moreover, using greedy decoding in Eq.\eqref{eq_likelihood}, the reconstructed results achieve high BLEU and ROUGE compared to the original model outputs. 
\begin{table}[t]
\small
\centering
\caption{KL divergence between the token distributions of surrogate model $\mathcal{M}_{\text{like}}$ and Qwen3-8B.}
\label{tab:kl_divergence}
\begin{tabular}{lcc}
\toprule
\textbf{Dataset} (held-out) & GSM8K & MATH-500 \\
\midrule
Qwen3-8B & 1.61 & 1.57 \\
\bottomrule
\end{tabular}
\vspace{-12pt}
\end{table}

A more direct evidence is presented in Table~\ref{tab:kl_divergence}, where we report the KL divergence between the token distributions of $\mathcal{M}_{\text{like}}$ and an open-source LLM (Qwen3), computed on responses sampled from Qwen3 over two held-out datasets, GSM8K and MATH-500. Considering the scale gap ($\approx 16$:$1$) between the original LLM and our surrogate model, the KL divergence is quite low, indicating that $\mathcal{M}_{\text{like}}$ closely replicates the generation distribution of LLMs, justifying its use as a proxy for approximating model activation.
% \begin{figure}[t]
% \centering
% %\setlength{\abovecaptionskip}{2pt}
% \includegraphics[width=0.95\linewidth]{figures/relative_strength_all_models_ordered.pdf}
% \caption{Average activation strength relative to root mind for each cognitive ability.}
% \label{figure_relative}
% %\vspace{-15pt}
% \end{figure}
\paragraph{\textbf{Pareto principle between latent mind dimensions and cognitive abilities across all LLMs.}} 
%Treating each root problem as a case, 
We then compare the latent minds induced by different cognitive variants and examine which dimensions are activated compared with the root problem's latent mind, hereafter called \emph{root mind}. 

The results reveal a Pareto principle, with the majority of dimensions forming a shared reasoning core and a small subset encoding ability-specific signatures. Specifically, any pair of cognitive variants exhibits a 82\%-97\% overlap of activated dimensions\footnote{This ratio is stable with $K \in \{128, 256, 512, 1024\}$, which confirms the robustness of our conclusion. To verify this is a reasoning-specific signature rather than a generic phenomenon of text processing, we provide a comparison in Appendix~\ref{append:pareto} against the random expectation (3.57\%) and a general dataset (WikiText-103, 53.3\%)}. The remaining non-overlapping dimensions, when treated as an ability classification task, achieve a precision of 0.72, indicating a strong alignment with their corresponding cognitive abilities.
%\footnote{In \emph{Appendix XX}, we conduct causal interventions to verify this functional mapping.}. 
Here, we do not focus on the specific dimension indices, as they may vary across training runs. Instead, we emphasize that these dimensions are largely shared across different LLMs, indicating that the same cognitive ability is well clustered in our framework regardless of model family.

\paragraph{\textbf{Cognitive variants amplify the scale of latent minds.}} 
In Figure~\ref{fig:combined_analysis}(b), we measure the activation strength of latent minds induced by different cognitive variants relative to the root mind. We first observe that the magnitude of amplification varies systematically across variants. Those involving minor perturbations (e.g., variants 4 and 5) show minimal changes, while variant 3 produces the strongest amplification. We argue that this pattern reflects the intrinsic difficulty of the reasoning process. In particular, when essential conditions are omitted in variant 3, the problem becomes ill-posed, forcing the model to explore a broader space of hypotheses, which is analogous to heightened cognitive effort in human problem solving under incomplete information~\cite{sweller1988cognitive}.
% and thereby substantially increasing reasoning intensity. This is 

Second, among these LLMs, DeepSeek-V3.2 has the highest activation across latent dimensions. This suggests a higher degree of \textbf{cognitive divergence} in DeepSeek, where stronger minds are required to encode and process information. This phenomenon indicates that different model architectures may employ varying bandwidths of latent mind when grappling with cognitive uncertainty.

% \begin{figure}[t]
% \centering
% %\setlength{\abovecaptionskip}{2pt}
% \includegraphics[width=0.8\linewidth]{figures/heatmap.pdf}
% \caption{Illustration of our {\ModelName} framework.}
% \label{figure_heapmap}
% %\vspace{-15pt}
% \end{figure}
\paragraph{\textbf{Latent minds capture different reasoning patterns.}}
For each cognitive variant, its reasoning pattern is typically reflected in the LLM-generated solutions. We extract the latent mind trajectories within these solutions to examine whether the latent minds can distinguish between cognitive variants. Specifically, since different data may contain varying numbers of steps, we employ Dynamic Time Warping (DTW)~\cite{berndt1994using} to compute an overall distance between two latent mind trajectories. The distances are then averaged over all solution instances for each variant pair.

Figure~\ref{fig:combined_analysis}(c) presents the min–max normalized distances among cognitive variants, where the root problem is assigned index 0. We can clearly observe that the latent minds of the root problem are closest to those of variant~1 and variant~4. In contrast, the distance between variant~3 and variant~9 is the largest. This phenomenon aligns well with the nature of their solution patterns. Variant~1 is a paraphrasing of the root problem, while variant~4 introduces irrelevant conditions without altering the core reasoning objective. As a result, the reasoning patterns required for solving these problems remains essentially unchanged. Conversely, variant~3 removes a necessary condition and trying to solve unsolvable problems, while variant~9 requires reverse reasoning with a fundamentally different objective. These two therefore induce latent minds that deviate most significantly.
% \begin{figure}[t]
% \centering
% %\setlength{\abovecaptionskip}{2pt}
% \includegraphics[width=0.9\linewidth]{figures/wrong_ratio.pdf}
% \caption{Average activation strength ratio of dimensions $\{3576, 11211, 12000\}$ in incorrect versus correct cases across different cognitive variants.}
% \label{figure_wrong_ratio}
% %\vspace{-15pt}
% \end{figure}
\begin{table*}[!ht]
\centering
\small
\setlength{\tabcolsep}{8pt}
\begin{tabular}{l|l|l|llll|l}
\toprule
\textbf{Model} & \textbf{Method} & \textbf{API Calls} & \textbf{GSM8K} & \textbf{MATH-500} & \textbf{AIME24} & \textbf{AIME25} & \textbf{Average} \\
\midrule

\multirow{5}{*}{DeepSeek-V3.2}
 & Standard & $\mathcal{O}(1)$
 & 97.0 & 96.4 & 90.7 & 91.3 & 93.9 \\
 & -SC & $\mathcal{O}(N)$
 & 97.4~{\scriptsize\textcolor{ForestGreen}{↑0.4}}
 & 97.2~{\scriptsize\textcolor{ForestGreen}{↑0.8}}
 & 93.3~{\scriptsize\textcolor{ForestGreen}{↑2.6}}
 & 93.3~{\scriptsize\textcolor{ForestGreen}{↑2.0}}
 & 95.3~{\scriptsize\textcolor{ForestGreen}{↑1.4}} \\
 & -LR & $\mathcal{O}(N)$ & 97.2~{\scriptsize\textcolor{ForestGreen}{↑0.2}}& 97.6~{\scriptsize\textcolor{ForestGreen}{↑1.2}} & 93.3~{\scriptsize\textcolor{ForestGreen}{↑2.6}}& 93.3~{\scriptsize\textcolor{ForestGreen}{↑2.0}}& 95.4~{\scriptsize\textcolor{ForestGreen}{↑1.5}}\\
 & -SV & $\mathcal{O}(NL)$
 & 97.5~{\scriptsize\textcolor{ForestGreen}{↑0.5}}
 & 97.2~{\scriptsize\textcolor{ForestGreen}{↑0.8}}
 & 94.0~{\scriptsize\textcolor{ForestGreen}{↑3.3}}
 & 93.3~{\scriptsize\textcolor{ForestGreen}{↑2.0}}
 & 95.5~{\scriptsize\textcolor{ForestGreen}{↑1.6}} \\

 & \cellcolor{gray!15}-\textbf{Ours} & \cellcolor{gray!15}$\mathcal{O}(N)$
 & \cellcolor{gray!15}97.8~{\scriptsize\textcolor{ForestGreen}{↑0.8}}
 & \cellcolor{gray!15}97.0~{\scriptsize\textcolor{ForestGreen}{↑0.6}}
 & \cellcolor{gray!15}96.7~{\scriptsize\textcolor{ForestGreen}{↑6.0}}
 & \cellcolor{gray!15}93.3~{\scriptsize\textcolor{ForestGreen}{↑2.0}}
 & \cellcolor{gray!15}\textbf{96.2}~{\scriptsize\textcolor{ForestGreen}{↑2.3}} \\

\midrule

\multirow{5}{*}{GPT-4o-0513}
 & Standard & $\mathcal{O}(1)$
 & 95.4 & 73.8 & 8.0 & 6.7 & 46.0 \\
 & -SC & $\mathcal{O}(N)$
 & 96.6~{\scriptsize\textcolor{ForestGreen}{↑1.2}}
 & 78.8~{\scriptsize\textcolor{ForestGreen}{↑5.0}}
 & 11.3~{\scriptsize\textcolor{ForestGreen}{↑3.3}}
 & 10.0~{\scriptsize\textcolor{ForestGreen}{↑3.3}}
 & 49.2~{\scriptsize\textcolor{ForestGreen}{↑3.2}} \\
 & -LR & $\mathcal{O}(N)$ & 97.1~{\scriptsize\textcolor{ForestGreen}{↑1.7}} & 78.8~{\scriptsize\textcolor{ForestGreen}{↑5.0}} & 13.3~{\scriptsize\textcolor{ForestGreen}{↑5.3}} & 13.3~{\scriptsize\textcolor{ForestGreen}{↑6.7}} & 50.6~{\scriptsize\textcolor{ForestGreen}{↑4.6}}\\
 & -SV & $\mathcal{O}(NL)$
 & 96.8~{\scriptsize\textcolor{ForestGreen}{↑1.4}}
 & 78.8~{\scriptsize\textcolor{ForestGreen}{↑5.0}}
 & 16.7~{\scriptsize\textcolor{ForestGreen}{↑8.7}}
 & 13.3~{\scriptsize\textcolor{ForestGreen}{↑6.6}}
 & 51.4~{\scriptsize\textcolor{ForestGreen}{↑5.4}} \\

 & \cellcolor{gray!15}-\textbf{Ours} & \cellcolor{gray!15}$\mathcal{O}(N)$
 & \cellcolor{gray!15}97.2~{\scriptsize\textcolor{ForestGreen}{↑1.8}}
 & \cellcolor{gray!15}80.2~{\scriptsize\textcolor{ForestGreen}{↑6.4}}
 & \cellcolor{gray!15}16.7~{\scriptsize\textcolor{ForestGreen}{↑8.7}}
 & \cellcolor{gray!15}20.0~{\scriptsize\textcolor{ForestGreen}{↑13.3}}
 & \cellcolor{gray!15}\textbf{53.5}~{\scriptsize\textcolor{ForestGreen}{↑7.5}} \\

\midrule

\multirow{7}{*}{Qwen3-8B}
 & Standard & $\mathcal{O}(1)$
 & 92.1 & 83.0 & 26.7 & 25.3 & 56.8 \\
 & -SC & $\mathcal{O}(N)$
 & 92.5~{\scriptsize\textcolor{ForestGreen}{↑0.4}}
 & 87.8~{\scriptsize\textcolor{ForestGreen}{↑4.8}}
 & 36.7~{\scriptsize\textcolor{ForestGreen}{↑10.0}}
 & 26.7~{\scriptsize\textcolor{ForestGreen}{↑1.4}}
 & 60.9~{\scriptsize\textcolor{ForestGreen}{↑4.1}} \\
 & -LR & $\mathcal{O}(N)$ & 95.8~{\scriptsize\textcolor{ForestGreen}{↑3.7}}& 88.0~{\scriptsize\textcolor{ForestGreen}{↑5.0}} & 26.7~{\scriptsize\textcolor{ForestGreen}{↑0.0}}& 26.7~{\scriptsize\textcolor{ForestGreen}{↑1.4}} & 59.3~{\scriptsize\textcolor{ForestGreen}{↑2.5}}\\
 & -PR & $\mathcal{O}(N)$ & 95.9~{\scriptsize\textcolor{ForestGreen}{↑3.8}}& 87.6~{\scriptsize\textcolor{ForestGreen}{↑4.6}} & 30.0~{\scriptsize\textcolor{ForestGreen}{↑3.3}} & 26.7~{\scriptsize\textcolor{ForestGreen}{↑1.4}} & 60.1~{\scriptsize\textcolor{ForestGreen}{↑3.3}} \\
 & -SV & $\mathcal{O}(NL)$
 & 93.3~{\scriptsize\textcolor{ForestGreen}{↑1.2}}
 & 86.6~{\scriptsize\textcolor{ForestGreen}{↑3.6}}
 & 33.3~{\scriptsize\textcolor{ForestGreen}{↑6.6}}
 & 23.3~{\scriptsize\textcolor{red}{↓2.0}}
 & 59.1~{\scriptsize\textcolor{ForestGreen}{↑2.3}} \\
 & -SI & $\mathcal{O}(N)$ & 96.0~{\scriptsize\textcolor{ForestGreen}{↑3.9}} & 87.2~{\scriptsize\textcolor{ForestGreen}{↑4.2}} & 33.3~{\scriptsize\textcolor{ForestGreen}{↑6.6}} & 26.7~{\scriptsize\textcolor{ForestGreen}{↑1.4}} & 60.8~{\scriptsize\textcolor{ForestGreen}{↑4.0}}\\
 & \cellcolor{gray!15}-\textbf{Ours} & \cellcolor{gray!15}$\mathcal{O}(N)$
 & \cellcolor{gray!15}96.2~{\scriptsize\textcolor{ForestGreen}{↑4.1}}
 & \cellcolor{gray!15}87.6~{\scriptsize\textcolor{ForestGreen}{↑4.6}}
 & \cellcolor{gray!15}36.7~{\scriptsize\textcolor{ForestGreen}{↑10.0}}
 & \cellcolor{gray!15}26.7~{\scriptsize\textcolor{ForestGreen}{↑1.4}}
 & \cellcolor{gray!15}\textbf{61.8}~{\scriptsize\textcolor{ForestGreen}{↑5.0}} \\

\midrule

\multirow{7}{*}{LLaMA-3.1-8B}
 & Standard & $\mathcal{O}(1)$
 & 85.1 & 47.2 & 3.3 & 1.3 & 34.2 \\
 & -SC & $\mathcal{O}(N)$
 & 92.3~{\scriptsize\textcolor{ForestGreen}{↑7.2}}
 & 55.2~{\scriptsize\textcolor{ForestGreen}{↑8.0}}
 & 6.7~{\scriptsize\textcolor{ForestGreen}{↑3.4}}
 & 3.3~{\scriptsize\textcolor{ForestGreen}{↑2.0}}
 & 39.4~{\scriptsize\textcolor{ForestGreen}{↑5.2}} \\
 & -LR & $\mathcal{O}(N)$ & 92.7~{\scriptsize\textcolor{ForestGreen}{↑7.6}}& 54.8~{\scriptsize\textcolor{ForestGreen}{↑7.6}}& 6.7~{\scriptsize\textcolor{ForestGreen}{↑3.4}}& 0.0~{\scriptsize\textcolor{red}{↓1.3}}& 38.6~{\scriptsize\textcolor{ForestGreen}{↑4.4}}\\
 & -PR & $\mathcal{O}(N)$ & 92.1~{\scriptsize\textcolor{ForestGreen}{↑7.0}} & 56.6~{\scriptsize\textcolor{ForestGreen}{↑9.4}} & 6.7~{\scriptsize\textcolor{ForestGreen}{↑3.4}} & 0.0~{\scriptsize\textcolor{red}{↓1.3}} & 38.9~{\scriptsize\textcolor{ForestGreen}{↑4.7}}\\
 & -SV & $\mathcal{O}(NL)$
 & 92.0~{\scriptsize\textcolor{ForestGreen}{↑6.9}}
 & 57.8~{\scriptsize\textcolor{ForestGreen}{↑10.6}}
 & 10.0~{\scriptsize\textcolor{ForestGreen}{↑6.7}}
 & 0.7~{\scriptsize\textcolor{red}{↓0.6}}
 & 40.1~{\scriptsize\textcolor{ForestGreen}{↑5.9}} \\
 & -SI & $\mathcal{O}(N)$ & 93.7~{\scriptsize\textcolor{ForestGreen}{↑8.6}} & 57.6~{\scriptsize\textcolor{ForestGreen}{↑10.4}} & 6.7~{\scriptsize\textcolor{ForestGreen}{↑3.4}} & 3.3~{\scriptsize\textcolor{ForestGreen}{↑2.0}} & 40.3~{\scriptsize\textcolor{ForestGreen}{↑6.1}}\\
 & \cellcolor{gray!15}-\textbf{Ours} & \cellcolor{gray!15}$\mathcal{O}(N)$
 & \cellcolor{gray!15}93.7~{\scriptsize\textcolor{ForestGreen}{↑8.6}}
 & \cellcolor{gray!15}57.0~{\scriptsize\textcolor{ForestGreen}{↑9.8}}
 & \cellcolor{gray!15}10.0~{\scriptsize\textcolor{ForestGreen}{↑6.7}}
 & \cellcolor{gray!15}3.3~{\scriptsize\textcolor{ForestGreen}{↑2.0}}
 & \cellcolor{gray!15}\textbf{41.0}~{\scriptsize\textcolor{ForestGreen}{↑6.8}} \\

\bottomrule
\end{tabular}
\caption{Performance comparison of different LLMs under various reasoning strategies.}
\label{tab:main_results}
\vspace{-10pt}
\end{table*}
% \begin{table*}[!ht]
% \centering
% \small
% \setlength{\tabcolsep}{6pt}
% \begin{tabular}{llccccc}
% \toprule
% \textbf{Model} & \textbf{Method} & \textbf{GSM8K} & \textbf{MATH-500} & \textbf{AIME24} & \textbf{AIME25} & \textbf{Average} \\
% \midrule

% \multirow{3}{*}{DeepSeek-V3.2}
%  & Standard & 97.0 & 96.4 & 90.7 & 91.3 & 92.78 \\
%  & Self-Consistency($K=5$)    & 97.4 & 97.2 & 93.3 & 93.3 & 94.81 \\
%  & Self-Verification($K=5$)    & 97.5 & 97.2 & 94.0 & 93.3 & -- \\
%  & Ours($K=5$)     & 97.7 & 97.0    & 94.0    & 93.3    & --    \\
% \midrule

% \multirow{3}{*}{GPT-4o-0513}
%  & Standard & 95.4 & 73.8 & 8.0  & 6.7 & 45.97 \\
%  & Self-Consistency($K=5$)  & 96.6 & 78.8 & 11.3 & 10.0 & 49.10 \\
%  & Self-Verification($K=5$)    & 96.8 & 78.8 & 16.7 & 13.3 & -- \\
%  & Ours($K=5$)     & 97.0   & 79.4   & 13.3   & 20.0   & --    \\
% \midrule

% \multirow{3}{*}{Qwen3-8B}
%  & Standard & 92.1 & 83.0 & 26.7 & 25.3 & 56.83 \\
%  & Self-Consistency($K=5$)  & 92.5 & 87.8 & 36.7 & 26.7 & 60.93 \\
%  & Self-Verification($K=5$)    & 93.3 & 86.6 & 33.3 & 23.3 & -- \\
%  & Ours($K=5$)     & 96.2   & 87.6   & 36.7   & 26.7   & --    \\
% \midrule

% \multirow{4}{*}{LLaMA-3.1-8B-Instruct}
%  & Standard & 85.1 & 47.2 & 3.3 & 1.3 & 34.24 \\
%  & Self-Consistency($K=5$) & 92.3 & 55.2 & 6.7 & 3.3 & 39.38 \\
%  & Self-Verification($K=5$)    & 92.0 & 57.8 & 10.0 & 0.7 & -- \\
%  & Ours($K=5$)     & 93.1    & 57.0  & 10.0  & 3.3  & --    \\

% \bottomrule
% \end{tabular}
% \caption{Performance comparison of different LLMs under various inference strategies.}
% \label{tab:main_results}
% \end{table*}
\begin{figure}[t]
\centering
\includegraphics[width=0.88\linewidth]{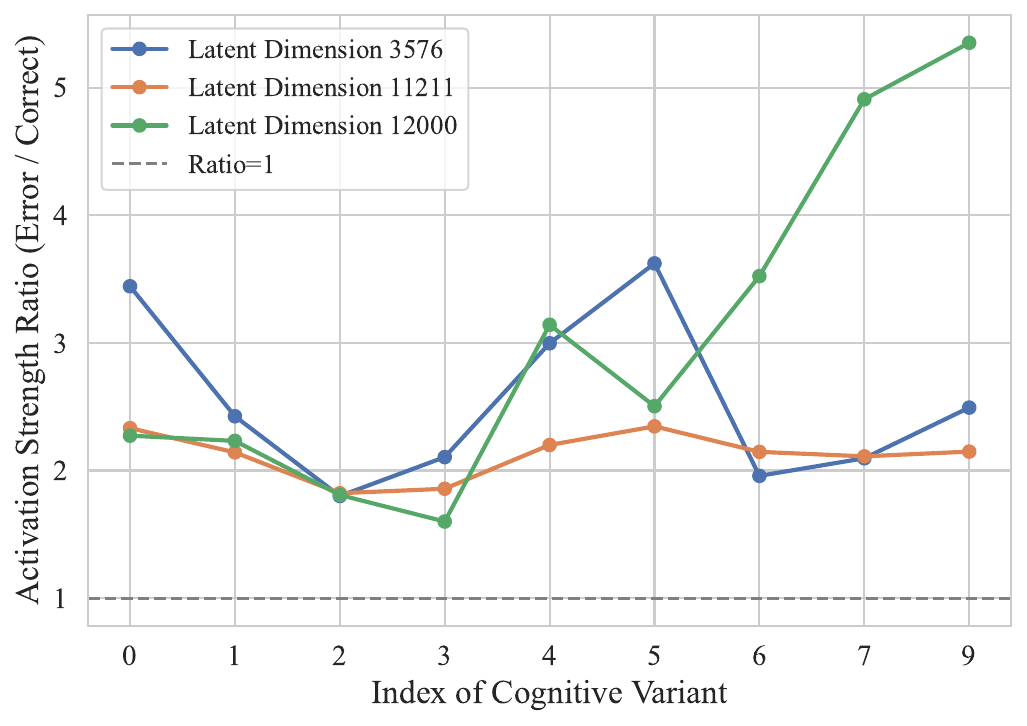}
\caption{Average activation strength ratio of dimensions $\{3576, 11211, 12000\}$ in incorrect versus correct cases across different cognitive variants.}
\label{figure_perplexity}
\vspace{-8pt}
\end{figure}
\paragraph{\textbf{Latent minds encode signals of reasoning correctness.}}
We further compare the activation of latent dimensions between correctly solved cases and erroneous ones. Specifically, for each case, we compute the activation ratio (i.e., the proportion of steps in which a dimension is activated) and the average activation strength for every latent dimension. We then analyze, dimension by dimension, the relative differences between correct and incorrect solutions. To reduce the influence of spurious activations, we only consider dimensions that are activated in more than 10\% of the erroneous cases.

Surprisingly, a pervasive trend emerges across all models. Latent activations are markedly intensified in erroneous cases. Specifically, the average activation strength across the entire latent space exhibits a 1.1× to 2.0× amplification when the model fails\footnote{In Appendix~\ref{append:solution_len}, we conduct a length-controlled analysis to show that this amplification is not caused by solution length.}. This suggests that erroneous reasoning is accompanied by an influx of stochastic noise or redundant signaling, which aligns with observations reported in prior studies~\cite{liu2025verifying}.  

Moreover, when examining the surrogate model, some dimensions stand out with markedly stronger activations in incorrect cases. As shown in Figure~\ref{figure_perplexity}, the average activation strength of dimensions $\{3576, 11211, 12000\}$ in erroneous cases is consistently more than twice that of correct cases across nearly all cognitive variants. Meanwhile, these dimensions tend to be activated 40\%–80\% less often in erroneous cases, despite having substantially higher activation magnitudes when they do activate. This pattern suggests that these dimensions may act as dominant but sporadic signals during failure modes, potentially amplifying noise or misleading reasoning trajectories.
% Overall, these results suggest that latent mind representations not only encode reasoning processes but also preserve structured, semantically meaningful relationships among distinct cognitive abilities.

\section{Enhancing LLM Reasoning via Latent Minds}

In this section, we leverage the insights obtained from our analysis to improve the reasoning performance of LLMs. Specifically, we aim to utilize the patterns of latent minds to identify and promote correct reasoning trajectories, thereby enhancing the overall reliability of LLM outputs.

\subsection{Latent-informed Candidate Prioritization}\label{sec:apply1}

Our application focuses on exploiting latent mind activations to rank multiple reasoning outputs. Given a problem $q$, we first sample $N$ candidate solutions $\{s_1, s_2, \dots, s_N\}$ from the target LLM $\mathcal{M}$. Then, for each candidate solution $s_i$, we extract its corresponding latent mind $\mathbf{z}_i$. Based on our previous findings, erroneous solutions tend to have higher activation magnitudes compared to correct solutions. Therefore, we define the average activation strength for solution $s_i$ as:

\begin{equation}
\label{eq:average}
A_i = \frac{1}{d} \sum_{j=1}^{d} |\mathbf{z}_{i,j}|.
\end{equation}

We rank the candidate solutions in ascending order of their activation strengths $A_i$. To further aggregate the reasoning results, we introduce a linear weighting strategy based on this ranking. Let $r_i \in \{1, \dots, N\}$ be the rank of solution $s_i$, where the solution with the highest activation strength (most likely to be incorrect) is assigned $r_i = 1$, and for each preceding position in the sorted rank, we increment the weight by $0.5$. The weight $w_i$ for solution $s_i$ is defined as: $w_i = 1 + 0.5 \times (r_i - 1)$.

For instance, with $N=5$, the weights assigned to solutions from lowest to highest activation strength would be $3.0, 2.5, 2.0, 1.5,$ and $1.0$, respectively. This approach prioritizes solutions that rely on fewer and more targeted latent dimensions, consistent with our observation that correct reasoning involves more focused activation patterns.

\subsection{Experimental Performance}
\paragraph{Evaluation Benchmark.}
We assess the effectiveness of our approach on four representative benchmarks:
\textbf{GSM8K}~\cite{cobbe2021training}: A basic dataset comprising 8.5K grade-school word problems. \textbf{MATH-500}~\cite{hendrycks2021measuring}: Consists of 500 high-school-level mathematics problems covering topics such as algebra, geometry, and basic calculus. \textbf{AIME 24}~\cite{aime24} and \textbf{AIME 25}~\cite{aime25}: Problems curated from the American Invitational Mathematics Examination (AIME), which challenge models with competition-level complexity and require highly sophisticated reasoning paths.
    %It is used to evaluate the model's generalization capabilities on novel, non-trivial problems that demand creative problem-solving heuristics.
\begin{figure*}[t]
\centering
\setlength{\abovecaptionskip}{2pt}
\includegraphics[width=\linewidth]{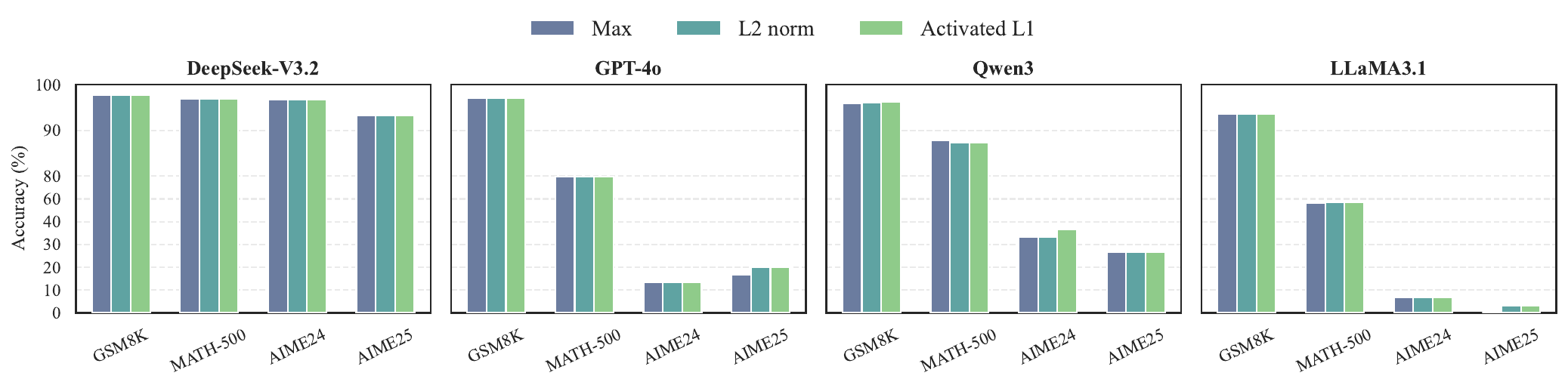}
\caption{Performance of different strategies for activation ranking.}
\label{figure_activation_method}
%\vspace{-15pt}
\end{figure*}
\paragraph{Baselines.}
Since our method focuses on ranking LLM-generated answers, we compare it with representative baselines that also introduce answer-ranking mechanisms: (1) \textbf{Standard}: The standard Chain-of-Thought prompting method. (2) \textbf{Self-Consistency (SC)}~\cite{wangself}: Samples $N$ reasoning paths and selects the most frequent answer via majority voting. (3) \textbf{Length-based Ranking (LR)}: Ranks candidate solutions based on their total token length, giving higher weights to longer solutions. (4) \textbf{Perplexity-based Ranking (PR)}: Ranks candidate solutions based on their perplexity. (5) \textbf{Self-Verification (SV)}~\cite{weng2023large}: For each candidate answer, construct a backward question and sample an LLM $L$ times to answer this question. The average correctness over these $L$ samples is used as to rank the candidate answers. To balance performance and computational cost, we set $L=5$ and employ LLaMA-3.1-8B for rewriting conclusions and GPT-4o-mini for answering. (6) \textbf{Self-Indicator (SI)}~\cite{liu2025verifying}: use the correlation matrix rank between the problem representations and answer representations as the score to rank answers. PR and SI are only used for open-source LLM backbones. 

\begin{figure*}[t]
\centering
\setlength{\abovecaptionskip}{0pt}
\begin{subfigure}
\centering
\includegraphics[width=\linewidth]{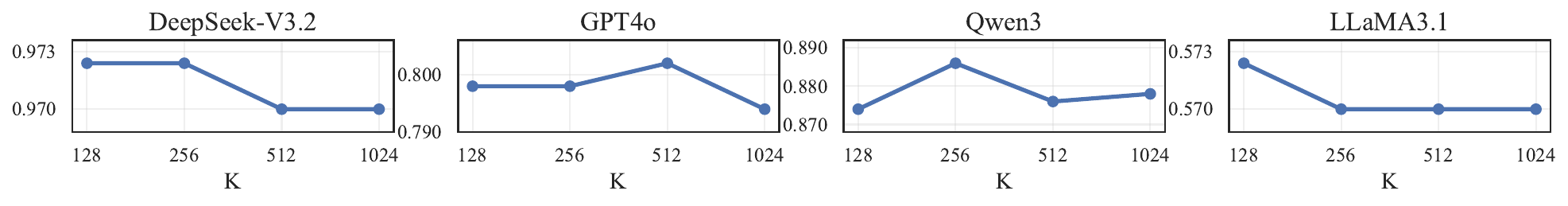}
\end{subfigure}
\\[-10pt] % 这里的间距非常容易控制
\begin{subfigure}
\centering
\includegraphics[width=\linewidth]{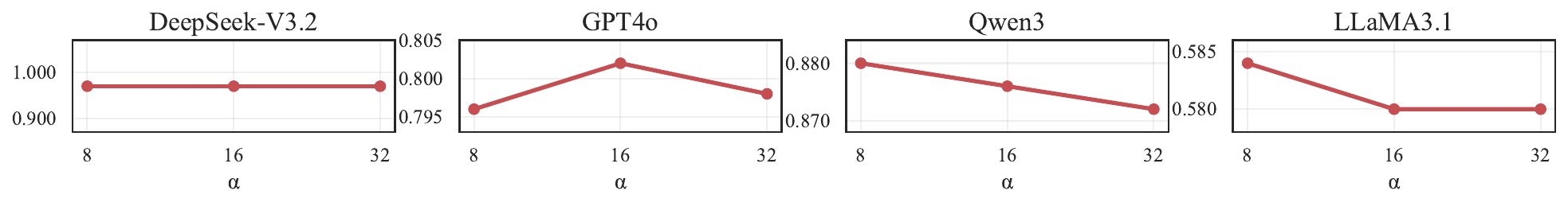}
\end{subfigure}\\[-5pt]
\caption{Performance with different values of $K$ and $d$ on MATH dataset.}
\label{fig:hyper}
\vspace{-10pt}
\end{figure*}

\subsection{Main Results}\label{section:main_result}
As shown in Table~\ref{tab:main_results}, our method consistently achieves the best or near-best performance across all benchmarks and backbone models. It yields substantial accuracy improvements over standard single-pass inference, with average gains of up to 7.5\%. Compared with SV, our method attains comparable or superior performance while requiring only $\mathcal{O}(N)$ API calls, avoiding the significant cost overhead. It also outperforms LR, demonstrating that our activation-based metric captures intrinsic reasoning quality rather than merely reflecting solution length. Notably, the improvements are more pronounced on challenging benchmarks such as AIME24 and AIME25, indicating that our approach is particularly effective at filtering erroneous reasoning paths in difficult problem settings. Furthermore, the consistent gains observed across both proprietary and open-source models demonstrate that our method is model-agnostic and generalizes well across different architectures.

\subsection{Analysis on Prioritization Strategies}
In Section~\ref{sec:apply1}, the prioritization strategy is to rank each solution $i$ based on the \emph{average} activation strength of its latent mind, as defined in Eq.~\eqref{eq:average}. Here, we investigate alternative strategies for calculating $A_i$, including \textit{Max} (taking the maximal activation in $\mathbf{z}_i$), \textit{L2 norm}, and \textit{Activated L1} (L1 norm over activated dimensions only).  

In Figure~\ref{figure_activation_method}, we observe that the differences between these strategies are relatively small. This indicates that the separation scale between correct and incorrect answers in the latent activation space is pronounced, making most reasonable aggregation methods similarly effective. In other words, the dominant signal corresponding to correct reasoning paths is sufficiently strong, so that simple aggregation heuristics like maximum selection or L1 norm yield comparable performance across different models and benchmarks.

\subsection{Analysis on Dimensions $K$ and $d$}\label{exp:hyper}
We investigate the impact of the number of activated dimensions $K \in \{128, 256, 512, 1024\}$ and the latent dimension $d = \alpha \times h$, where $h$ is the hidden dimension of the surrogate model and $\alpha \in \{8, 16, 32\}$. Due to space limit, we report the results on MATH dataset in Figure~\ref{fig:hyper} and provide other results in Appendix~\ref{append:addition_results}.

Overall, the performance degrades when either $K$ or $d$ is too small or excessively large, indicating that both dimensions need to be properly balanced. When they are too small, each latent mind is forced to encode a higher amount of information within a limited representational capacity.
As a result, the signals corresponding to correct and incorrect reasoning may be overwhelmed by other information. Conversely, when they become overly large, the signals tend to be excessively dispersed across latent dimensions, which weakens their influence after aggregation and leads to diminished performance.

\section{Conclusions}
In this paper, we proposed UniCog, a unified framework for analyzing LLM cognition through the lens of latent mind space. UniCog revealed how different cognitive abilities are engaged during reasoning, including its Pareto principle, ability-specific latent signatures, and systematic activation amplification effect. Moreover, we demonstrated that latent mind signals can be effectively leveraged to improve reasoning reliability in a plug-and-play manner. UniCog provides a cognitively grounded perspective on LLM reasoning and opens new directions for interpretability.
\section*{Limitations}

Despite the promising results of our {\ModelName} framework, some limitations remain to be addressed in future work.

A primary constraint of our study is the absence of standard disentanglement metrics, such as DCI~\cite{eastwood2018framework}, MIG~\cite{chen2018isolating}, or SAP~\cite{kumar2018variational}. Calculating these metrics requires a dataset with known, ground-truth cognitive labels for each reasoning step, which currently does not exist for complex mathematical reasoning. While our correlation analysis and cross-domain comparisons provide strong evidence for the functional role of the latent mind, we acknowledge that the precise degree of structural disentanglement in the latent space remains to be formally quantified as more granularly annotated datasets become available.

Our current evaluation is primarily focused on mathematical reasoning, a domain characterized by rigid logical structures and objective correctness. While it serves as a robust testbed for cognitive analysis, it remains unclear whether our findings generalize to more subjective or open-ended tasks, such as creative writing or commonsense reasoning. The effectiveness of using latent mind strength as a proxy for reasoning reliability may vary in domains where logical density is lower.

The current application of the latent mind is post-hoc, utilized as a prioritization strategy to select among pre-generated candidates. While this plug-and-play approach offers high efficiency and broad compatibility with both open-source and proprietary LLMs, it does not actively steer the model's internal state during the autoregressive generation process. Future research could explore the use of latent mind signals as a real-time feedback mechanism (e.g., via activation steering or constrained decoding) to proactively guide the model away from erroneous reasoning paths before they are fully materialized in language.

% \section*{Acknowledgments}

% This document has been adapted
% by Steven Bethard, Ryan Cotterell and Rui Yan
% from the instructions for earlier ACL and NAACL proceedings, including those for
% ACL 2019 by Douwe Kiela and Ivan Vuli\'{c},
% NAACL 2019 by Stephanie Lukin and Alla Roskovskaya,
% ACL 2018 by Shay Cohen, Kevin Gimpel, and Wei Lu,
% NAACL 2018 by Margaret Mitchell and Stephanie Lukin,
% Bib\TeX{} suggestions for (NA)ACL 2017/2018 from Jason Eisner,
% ACL 2017 by Dan Gildea and Min-Yen Kan,
% NAACL 2017 by Margaret Mitchell,
% ACL 2012 by Maggie Li and Michael White,
% ACL 2010 by Jing-Shin Chang and Philipp Koehn,
% ACL 2008 by Johanna D. Moore, Simone Teufel, James Allan, and Sadaoki Furui,
% ACL 2005 by Hwee Tou Ng and Kemal Oflazer,
% ACL 2002 by Eugene Charniak and Dekang Lin,
% and earlier ACL and EACL formats written by several people, including
% John Chen, Henry S. Thompson and Donald Walker.
% Additional elements were taken from the formatting instructions of the \emph{International Joint Conference on Artificial Intelligence} and the \emph{Conference on Computer Vision and Pattern Recognition}.

% Bibliography entries for the entire Anthology, followed by custom entries
%\bibliography{custom,anthology-overleaf-1,anthology-overleaf-2}

% Custom bibliography entries only
\bibliography{custom}

\appendix

\section{Implementation Details}
\label{append:implement}
In our framework, both the surrogate likelihood model $\mathcal{M}_{\text{like}}$ and the posterior model are implemented based on the Qwen2.5-0.5B architecture. The prior distribution $p(Z)$ is a fixed isotropic Gaussian distribution: 
\begin{equation}
Z \sim \mathcal{N}(0, \sigma^2 I), \quad \sigma = 0.01.
\end{equation}

\begin{figure}[t]
\centering
\includegraphics[width=\linewidth]{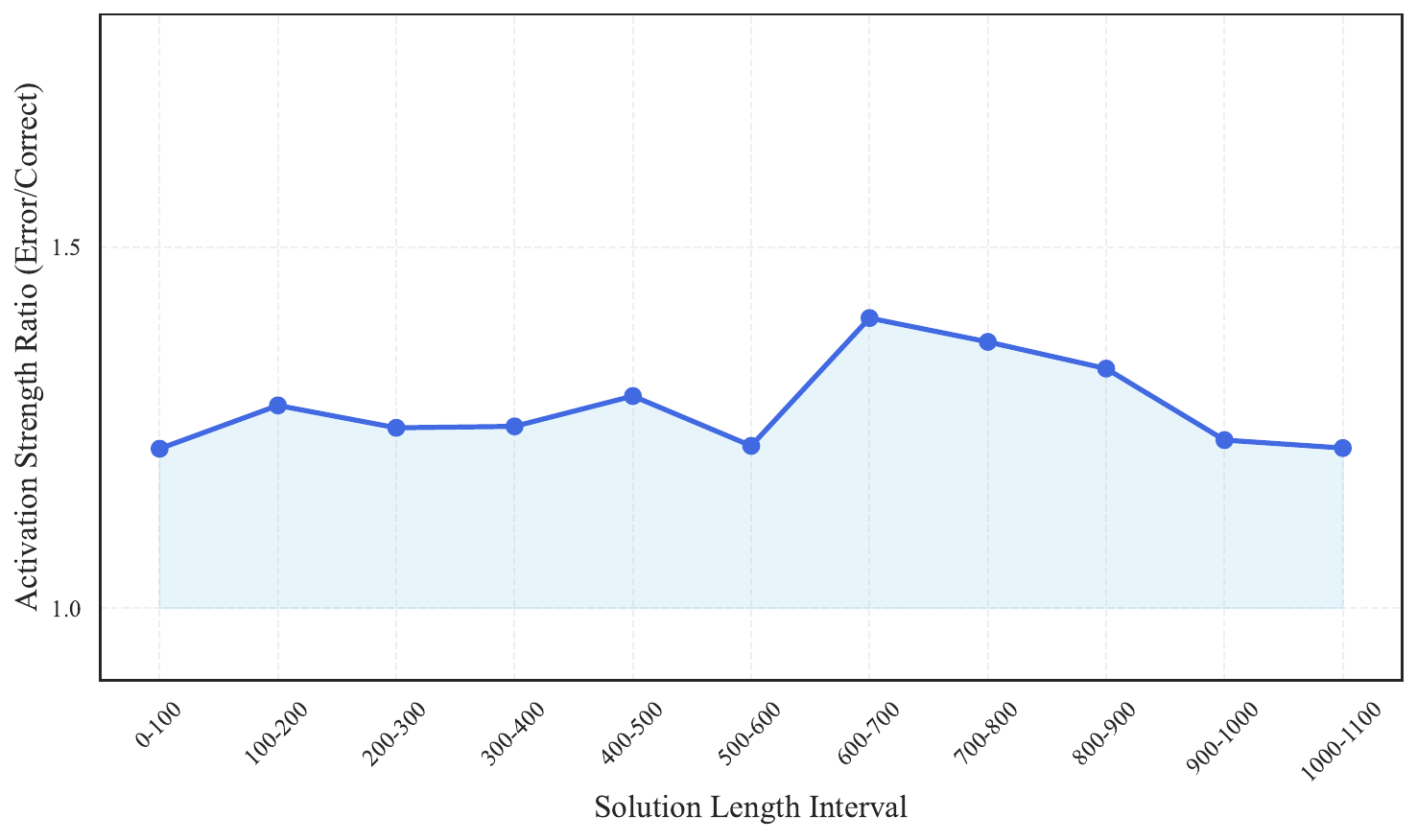}
\caption{Average activation ratio between incorrect and correct cases across length intervals.}
\label{figure_solution_length}
\vspace{-10pt}
\end{figure}
We define the latent mind space dimension as $d = \alpha \times h$, where $h = 896$ is the hidden dimension of Qwen2.5-0.5B. The expansion factor $\alpha = 16$, resulting in a total latent dimensionality of $d = 14,336$. In SAE setup, $K$ is set as $512$. 
\begin{figure*}[t]
\centering
\setlength{\abovecaptionskip}{0pt}
\begin{subfigure}
\centering
\includegraphics[width=\linewidth]{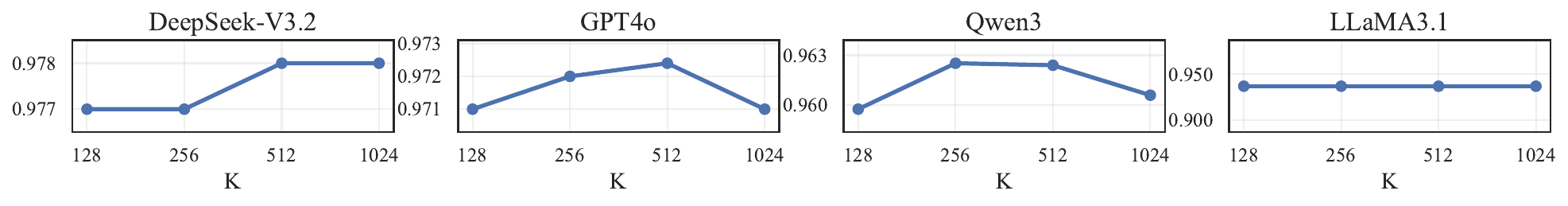}
\end{subfigure}
\\[-10pt] % 这里的间距非常容易控制
\begin{subfigure}
\centering
\includegraphics[width=\linewidth]{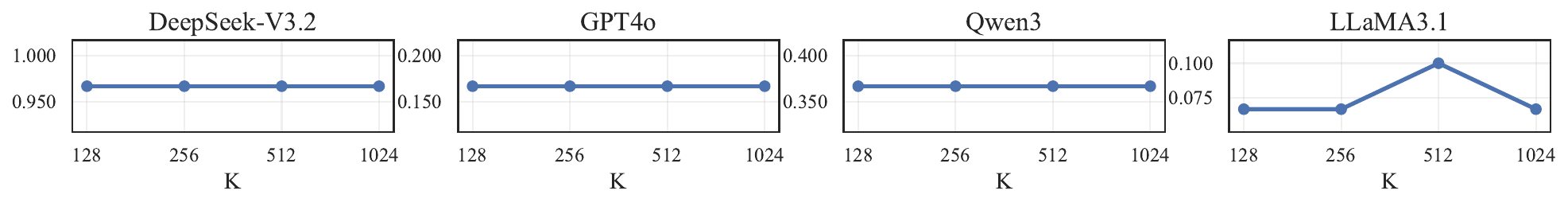}
\end{subfigure}
\\[-10pt]
\begin{subfigure}
\centering
\includegraphics[width=\linewidth]{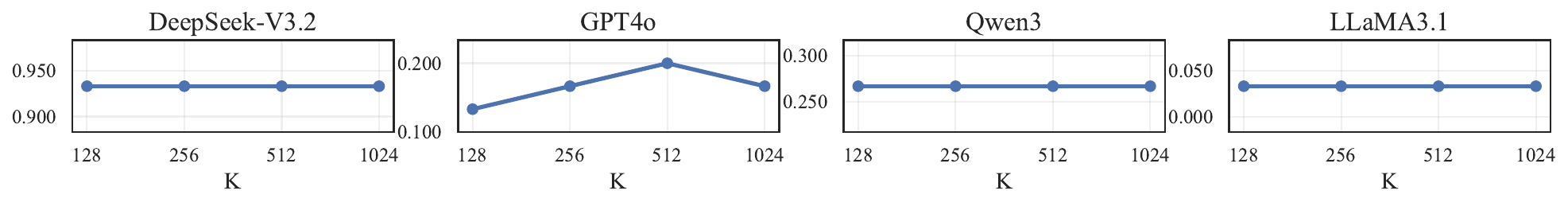}
\end{subfigure}\\[-5pt]
\caption{Performance with different values of $K$ on GSM8K, AIME24, and AIME25.}
\label{fig:hyper_K}
\vspace{-5pt}
\end{figure*}
\begin{figure*}[t]
\centering
\setlength{\abovecaptionskip}{0pt}
\begin{subfigure}
\centering
\includegraphics[width=\linewidth]{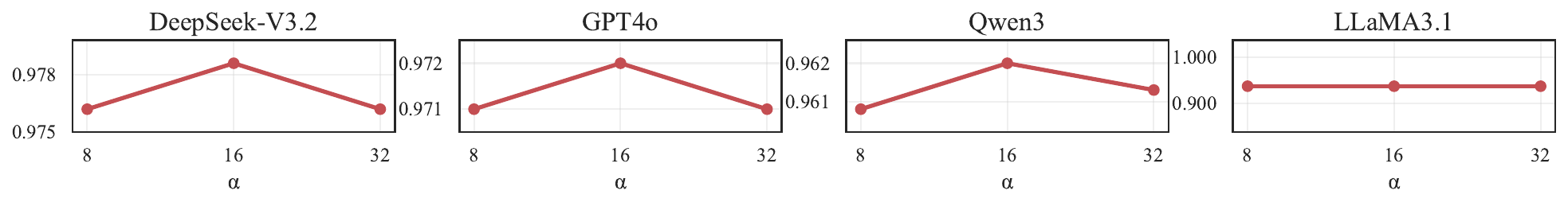}
\end{subfigure}
\\[-10pt] % 这里的间距非常容易控制
\begin{subfigure}
\centering
\includegraphics[width=\linewidth]{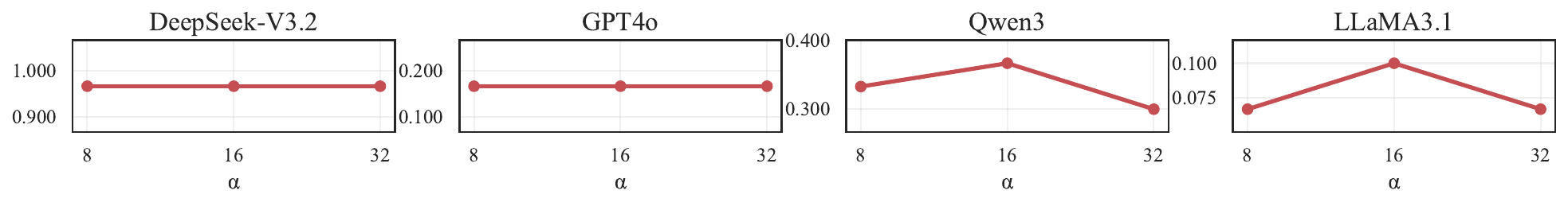}
\end{subfigure}
\\[-10pt]
\begin{subfigure}
\centering
\includegraphics[width=\linewidth]{figures/d_AIME24.pdf}
\end{subfigure}\\[-5pt]
\caption{Performance with different values of $d=\alpha \cdot h$ on GSM8K, AIME24, and AIME25.}
\label{fig:hyper_d}
\vspace{-5pt}
\end{figure*}

During preprocessing NuminaMath-CoT dataset, all sequences are partitioned into sentence-level units, which serve as $X$ for model training. Our {\ModelName} is trained using AdamW with a learning rate of $1e^{-5}$ and a batch size of $256$, along with linear weight decay. When sampling reasoning outputs from open-source LLMs, Qwen-3 is used in non-thinking mode with default temperature, and Llama-3.1-8B-Instruct is sampled with temperature $0.8$ and top-p $0.95$.

\section{Non-Triviality of the Pareto principle}\label{append:pareto}
To quantify the non-triviality of the observed Pareto principle in Section~\ref{section:finding}, we compared our results against a random-expectation baseline. For a total dimension $d=14,336$ and $K=512$, the expected random overlap ratio between any two subsets is only $\frac{K}{d}\approx 3.57\%$. Moreover, we further compare the latent minds on CogMath with those on a general language modeling dataset, WikiText-103~\cite{merity2016pointer}. We find that the overlap between their latent minds is only 53.3\%. This result demonstrates that the observed high-dimensional overlap is inherently coupled with mathematical reasoning processes, rather than being a generic byproduct of text processing. 

\section{Activation Strength Analysis across Solution Lengths}\label{append:solution_len}
In Section~\ref{section:finding}, we observed that, on CogMath, latent minds of LLMs exhibit higher activation in erroneous reasoning cases compared to correct cases. To verify that this amplification is not a byproduct of solution length, we further partition the dataset into consecutive length intervals of 100 tokens each. For each interval, we identify all error–correct pairs whose solution lengths fall within the same interval and computed the ratio of average activation in erroneous cases to that in correct cases.

As shown in Figure~\ref{figure_solution_length}, the error/correct activation ratio remains consistently above 1 across intervals, without a systematic increase for longer answers. Combined with the observations in Section~\ref{section:main_result}, where our method achieves higher reasoning accuracy compared to length-ranking based approaches, this confirms that the amplification phenomenon reflects intrinsic characteristics of erroneous reasoning rather than being caused by solution length.

\section{Additional Results on the impact on Dimensions $K$ and $d$}\label{append:addition_results}
Figure~\ref{fig:hyper_K} and Figure~\ref{fig:hyper_d} report additional results on GSM8K, AIME24, and AIME25 for different values of $K$ and $d$, respectively, complementing the MATH analysis in Section~\ref{exp:hyper}.

\end{document}